\documentclass[accepted, specialissue]{melba}

\usepackage{amsmath,amsfonts}

\usepackage{booktabs}
\usepackage{graphicx}
\usepackage{acronym}

\melbaid{2026:003}  
\doi{10.59275/j.melba.2026-a3eb}
\melbaauthors{Ammeling et al.}  
\email{jonas.ammeling@thi.de}
\volume{2026}
\firstpageno{38}  
\melbayear{2026}  
\datesubmitted{2025-07-15}  
\datepublished{2026-01-30}  

\melbaspecialissue{MELBA–BVM 2025 Special Issue}
\melbaspecialissueeditors{Andreas Maier, Thomas Deserno, Heinz Handels, Klaus Maier-Hein, Christian Palm, Thomas Tolxdorff, Katharina Breininger}

\ShortHeadings{Benchmarking Foundation Models for Mitotic Figure Classification}{Ammeling et al.}

\title{Benchmarking Foundation Models for Mitotic Figure Classification}

\author{
	\firstname Jonas \surname Ammeling\aff{1}\orcid{0000-0002-0335-1194},
	\name Jonathan Ganz\aff{2, 1}\orcid{0009-0008-1299-8716},
    \name Emely Rosbach\aff{1}\orcid{0009-0001-7526-9923},
    \name Ludwig Lausser\aff{1},
    \name Christof A. Bertram\aff{3}\orcid{0000-0002-2402-9997},
    \name Katharina Breininger\aff{4, 5}\orcid{0000-0001-7600-5869},
    \name Marc Aubreville\aff{6}\orcid{0000-0002-5294-5247}
}
\affiliations{
	\num 1 \addr Technische Hochschule Ingolstadt, AImotion, Ingolstadt, DE \\
        \num 2 \addr MIRA vision Microscopy GmbH, Göppingen, DE \\
	\num 3 \addr University of Veterinary Medicine, Institute of Pathology, Vienna, AU \\
	\num 4 \addr Center for AI and Data Science, Julius-Maximilians-Universität Würzburg, Würzburg, DE \\
    \num 5 \addr Department Artificial Intelligence in Biomedical Engineering, Friedrich-Alexander-Universität Erlangen-Nürnberg, Erlangen, DE \\
    \num 6 \addr Flensburg University of Applied Sciences, Flensburg, DE
}

\abstract{
    The performance of deep learning models is known to scale with data quantity and diversity. In pathology, as in many other medical imaging domains, the availability of labeled images for a specific task is often limited. Self-supervised learning techniques have enabled the use of vast amounts of unlabeled data to train large-scale neural networks, i.e., foundation models, that can address the limited data problem by providing semantically rich feature vectors that can generalize well to new tasks with minimal training effort increasing model performance and robustness. In this work, we investigate the use of foundation models for mitotic figure classification. The mitotic count, which can be derived from this classification task, is an independent prognostic marker for specific tumors and part of certain tumor grading systems. In particular, we investigate the data scaling laws on multiple current foundation models and evaluate their robustness to unseen tumor domains. Next to the commonly used linear probing paradigm, we also adapt the models using low-rank adaptation (LoRA) of their attention mechanisms. We compare all models against end-to-end-trained baselines, both CNNs and Vision Transformers. Our results demonstrate that LoRA-adapted foundation models provide superior performance to those adapted with standard linear probing, reaching performance levels close to 100 \% data availability with only 10 \% of training data. Furthermore, LoRA-adaptation of the most recent foundation models almost closes the out-of-domain performance gap when evaluated on unseen tumor domains. However, full fine-tuning of traditional architectures still yields competitive performance.}


\begin{document}

\twocolumn[\maketitle]

\newacro{tcga}[TCGA]{The Cancer Genome Atlas}
\newacro{wsi}[WSI]{whole slide image}
\newacro{ssl}[SSL]{self-supervised learning}
\newacro{vit}[ViT]{Vision Transformer}
\newacro{lora}[LoRA]{Low-Rank Adaptation}
\newacro{roi}[ROI]{region of interest}
\newacroplural{roi}[ROIs]{regions of interest}
\newacro{peft}[PEFT]{parameter-efficient fine-tuning}
\newacro{ccmct}[CCMCT]{canine cutaneous mast cell tumor}
\newacro{cmc}[CMC]{canine mammary carcinoma}

\section{Introduction}

\enluminure{S}{elf-supervised} learning (SSL) is transforming the landscape of publicly available methods for computational pathology \citep{Khan2024}. By leveraging vast amounts of unlabeled data, SSL overcomes the limitations of traditional supervised approaches, which are constrained by the availability of expert-annotated datasets. This is particularly advantageous in computational pathology, where annotation is a labor-intensive and time-consuming process requiring highly trained pathologists to examine large, high-resolution \acp{wsi} with diverse morphological structures across tissue types. The annotation process is further challenged by fatigue \citep{stec2018systematic} and cognitive biases ~\citep{Aeffner2017, Viray2013, Leiser2023}, leading to variability in label quality and high inter-rater variability \citep{Smits2014}. Additionally, differences in staining protocols and scanning devices across institutions can alter tissue appearance, complicating the development and deployment of robust methods \citep{Aubreville2024}. The scarcity of high-quality and large-scale datasets ultimately limits the advancement of supervised models and hinders their generalizability across diverse clinical settings. 
With the advent of \ac{ssl} techniques such as SimCLR \citep{Chen2020simclrv1, Chen2020simclrv2}, MoCo \citep{chen2020mocov1, Chen2020mocov2} and DINO \citep{Caron2021dinov1, 23dinov2}, large-scale neural networks were able to train on datasets with sizes beyond 1 billion images \citep{Goyal2021} surpassing the performance of models trained on labeled data on competitive benchmarks like ImageNet \citep{Tomasev2022, chen2020mocov1, Deng2009}. These models, often large \acp{vit} \citep{Dosovitskiy2020}, are commonly referred to as foundation models due to their ability to adapt to a wide range of downstream tasks with little to no fine-tuning. Because of their comprehensive pretraining, they are capable of generating semantically rich embeddings, which enables them to perform well in tasks such as few-shot learning and to serve as a robust foundation for a multitude of downstream applications \citep{Zhang2024-LAF, Zhang2024-CPF}.

Computational pathology is especially well-suited for \ac{ssl}, as it routinely generates large volumes of image data. Public resources such as \ac{tcga} \citep{Tomczak2015} provide access to tens of thousands of \acp{wsi} from multiple institutions, offering a diverse and abundant source of training data. Leveraging these resources, several pathology-specific foundation models have recently been developed using \ac{tcga} \citep{Chen2022, Wang2022-TUC, Kang2022, Filiot2023} and other public or proprietary datasets \citep{Chen2024UNI, Vorontsov2023, Zimmermann2024, Filiot2024Phikonv2, Saillard2024, Xu2024}. These models are typically evaluated on downstream tasks such as tumor subtyping, tissue classification and mutation prediction.

A recent work by \cite{Vorontsov2023} has shown that the image embeddings of foundation models can also be utilized effectively for tasks that require fine-grained morphological features, like mitotic figure classification. This task is of particular importance in computational pathology, as accurate identification and classification of mitotic figures are crucial for assessing the aggressiveness of tumors and estimating the outcome of tumor patients (prognostication) \citep{ELSTON1991}. However, the detection of mitotic figures remains challenging due to their morphological similarity to other cellular structures \citep{Donovan2021} and their sparse occurrence within tissue sections \citep{Aubreville2020most}. In modern detection pipelines mitotic figure classification is often employed as a second-stage process, following the initial identification of candidate objects \citep{li2018deepmitosis,Aubreville2024}. The integration of foundation models into these pipelines holds promise for improving classification performance or to reduce the need for large amounts of annotated data. 
Foundation models are typically adapted to new downstream tasks, such as mitotic figure classification, using techniques such as linear probing or model adaptation techniques like \ac{lora} \citep{Hu2022}. Linear probing involves training a simple linear classifier on top of the representations produced by the frozen foundation model, providing a fast and computationally efficient way to assess the quality of learned features. In contrast, adaptation techniques employ methods that selectively fine-tune parts of the foundation model to better tailor its representations to the downstream task. For example, \ac{lora} introduces trainable low-rank matrices into selected layers of the model, enabling more flexible fine-tuning with minimal changes to the original model parameters. Both approaches can reduce the need for extensive fine-tuning and large annotated datasets, and the limited number of trainable parameters provides a regularizing effect that helps prevent overfitting.

However, there has been no in-depth analysis of how the performance of such classifiers depends on the size of the training set or how robust they are to domains shifts arising from differences between source and target image characteristics. This work, which extends previous work published as a conference paper \citep{Ganz2025bvm}, aims to address these gaps by systematically investigating the scaling laws of several state-of-the-art pathology-specific foundation models for mitotic figure classification. To provide a comprehensive evaluation, we benchmark foundation model-based classifiers against several baseline methods across two publicly available mitotic figure datasets and evaluate the impact of the training set size using both linear probing and \ac{lora}.


\section{Related Work}

The following section outlines the foundational principles and recent developments in self-supervised learning that underpin many state-of-the-art models in computational pathology.

\subsection{Self-Supervised Learning}
The development of \ac{ssl} techniques marked a paradigm shift by enabling the training of large-scale neural networks on massive unlabeled datasets. In comparison to supervised learning techniques, where a model is trained on a specific task based on the available labeled data, \ac{ssl} learns generic representations useful across many tasks without any labels by utilizing the intrinsic structure of the data. A major branch of \ac{ssl} methods in computer vision is built based upon contrastive learning, which aims to create an embedding space where data points are organized based on their assumed similarity. SimCLR \citep{Chen2020simclrv1, Chen2020simclrv2} and MoCo \citep{chen2020mocov1, Chen2020mocov2} are the most prominent methods in this paradigm, where two views of the same image, slightly altered by standard augmentation techniques, such as changing the color or cropping, are to be mapped to similar representations, whereas representations from views of different images are pushed further apart in latent space. 
Another branch of \ac{ssl} methods is based on self-distillation, such as DINO \citep{Caron2021dinov1}, where a teacher-student framework is employed. In this setup, both the teacher and student networks share the same architecture but receive differently augmented views of the same image. The student, who only sees a smaller image crop, is trained to match the output distribution of the teacher, who sees a larger image crop. The distillation mechanism in this setup is given by the teacher being updated as an exponential moving average of the student’s parameters. Another self-distillation method that builds on DINO is iBOT \citep{Zhou2021}, where a masked image modeling objective is added that is applied in the latent space directly, such that the target reconstruction is not the original image pixels but the same patches embedded through the teacher network. DINOv2 \citep{23dinov2} further builds on iBOT and is used by the majority of the recently published foundation models \cite{Chen2023UNI, Vorontsov2023}. They improved the performance by modifying the training recipe and the architecture with better hyperparameter and regularizer such as KoLeo \citep{Sablayrolles2018} to be more effective and stable at larger model and data sizes.

Due to the ability to leverage large-scale, unlabeled datasets, \ac{ssl} techniques have gained a lot of attention across many healthcare applications \citep{Khan2024}, where labels are costly and time-consuming to acquire, such as clinical language models \citep{Lee2019, Chen2023, Yang2022scBert, Singhal2023}, medical image analysis \citep{Ma2024, Chen2024UNI, Wang2024, Xu2024, Alber2025, Filiot2024Phikonv2, Dippel2024}, vision and language applications \citep{Zhang2020, Wang2022-MedCLIP, Huang2023, Lu2024, Ahmed2024, Chen2024SlideChat}, and omics research \citep{Yang2022GatorTron, Celaj2023,Zhou2023}. 

\subsection{Foundation Models in Pathology}
Several pathology specific foundation models were published recently with promising performance across a multitude of downstream applications \citep{Ciga2022, Wang2022-TUC, Xu2024, Alber2025, Filiot2025, Saillard2024, Filiot2024Phikonv2, Dippel2024, Zimmermann2024}, with mitotic figure classification being included by some of these works \citep{Wang2022-TUC, Vorontsov2023,Zimmermann2024,Shen2024}.  In particular, \cite{Wang2022-TUC} proposed SRCL, an \ac{ssl} method based on MoCov3 \citep{Chen2021} along with CTransPath, a model architecture that combines convolutional layers with the Swin Transfomer model \citep{Liu2021}. Besides typical downstream tasks such as tile-level and slide-level classification, they also evaluated mitotic figure detection on the MIDOG 2021 \citep{Aubreville2023midog} dataset reporting an F1 score of $0.7332$ on a custom test split, showing superior performance of SRCL to other SSL frameworks such as SimCLR and DINO \citep{Wang2022-TUC}. They used the pre-trained CTransPath encoder as the backbone for the Faster R-CNN framework and performed full fine-tuning to adapt to the downstream task.  

\cite{Vorontsov2023} introduced Virchow, a ViT-H model trained with DINOv2 \citep{23dinov2} on a massive proprietary dataset consisting of 2 billion tiles from almost $1.5$ million slides across 17 tissue types. One of their downstream evaluation included mitotic figure classification on the MIDOG++ dataset \citep{Aubreville2023midogpp}. They extracted patches of size $224\times224$ for each annotation from the original \acp{roi} and performed linear probing to train a classifier to distinguish between patches of mitotic figures and non-mitotic figures. They report an F1 score of $0.787$ on a custom test split, outperforming all other tested foundation models. 

Building on Virchow, \cite{Zimmermann2024} released Virchow2 based on ViT-H and Virchow2G based on ViT-G, both trained with DINOv2 \citep{23dinov2} on $1.7$ billion and $1.9$ billion tiles, respectively, from $3.1$ million proprietary slides. Compared to the original Virchow model, they refined the training recipe to better suit pathology applications, incorporating mixed magnification training and exploring the effects of increased model and data scale as well as greater data diversity. They evaluated mitotic figure classification on the MIDOG++ dataset as well using the same linear probing protocol and test split as in the original Virchow work and report improved F1 scores of $0.804$ for Virchow2 and $0.836$ for Virchow2G.

\cite{Shen2024} introduced the Optimised Mitoses Generator Network (OMG-Net) to perform mitotic figure detection. Their 2-stage framework utilized SAM \citep{Kirillov2023}, a promptable foundation model with zero-shot capabilities to transfer to new image distributions and tasks, as first stage to outline candidate cells, followed by an adapted ResNet18 \citep{He2015} that distinguishes mitotic figures. They combined publicly available mitotic figure datasets such as ICPR \citep{Ludovic2013}, TUPAC \citep{Veta2019}, MIDOG++ \citep{Aubreville2023midogpp}, two fully annotated \ac{wsi} datasets for \ac{ccmct} \citep{Bertram2019} and for \ac{cmc} \citep{Aubreville2020}, together with an in-house dataset of human soft tissue tumor (STT) to create a large database with $74620$ mitotic figures to train their pipeline. They report F1 scores on a MIDOG++ test split ranging between $0.64$ on neuroendocrine tumors to $0.86$ for cutaneous mast cell tumors.
%

\cite{Xu2024} introduced Prov-GigaPath, a new foundation model for slide-level pretraining. They first trained a ViT-G architecture on tile-level using DINOv2, followed by slide-level pretraining using a masked autoencoder \citep{He2021} and LongNet \citep{Ding2023} to scale to thousands of image-tiles. They trained their model on $1.3$ billion tiles from $171,189$ proprietary slides from Providence Health and Services. Prov-GigaPath was evaluated on 17 genomic prediction tasks and 9 cancer subtyping tasks using both Providence and TCGA data. 

\cite{Chen2024UNI} introduced UNI, a tile-level foundation model based on ViT-L, trained with DINOv2 on more than 100 million tiles from 100K proprietary slides. They evaluated UNI on 34 clinical tasks with varying diagnostic difficulty, such as nuclear segmentation, primary and metastatic cancer detection, cancer grading and subtyping, molecular subtyping and several pan-cancer classification tasks. 

\cite{Filiot2023} introduced Phikon, a ViT-B model trained with iBOT \citep{Zhou2021}, combining masked image modeling and contrastive learning. They trained their model on $43.3$ million tiles from 6093 TCGA slides. They evaluated their performance across 17 downstream tasks including tile-level and slide-level tasks such as subtype classification, genomic alterations, and survival prediction. 

\cite{Saillard2024} introduced H-optimus-0, a ViT-G model trained with DINOv2 on more than 500K proprietary slides. There is no exact number of tiles on which they trained their model but they mentioned more than several 100 million tiles. They evaluated across a wide range of downstream tasks as well covering tasks such as tissue classification, mutation prediction, and survival analysis. 

\begin{figure*}[!ht]
    \centering
    \includegraphics[width=0.8\textwidth]{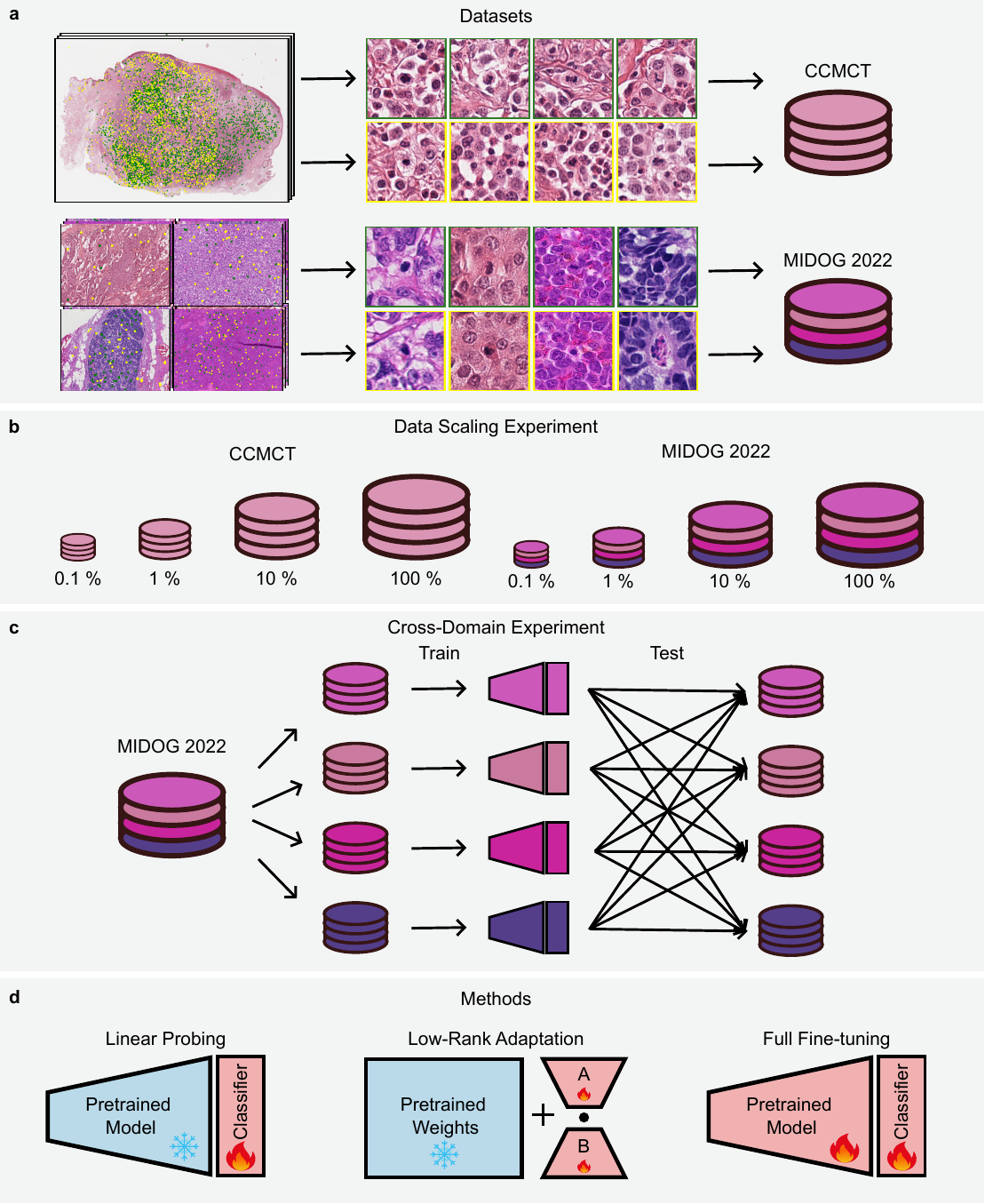}
    \caption{Benchmark study overview. \textbf{a)} Exemplary overview of datasets. Green shows mitotic figures and yellow shows hard negatives. During inference we extract patches of size $224\times224$ around these annotations for evaluation. \textbf{b)} Overview of dataset scaling experiments. \textbf{c)} Schematic overview of the cross-domain experiment. We train a model on each domain separately and evaluate across all domains. \textbf{d)} Overview of evaluated methods.}
    \label{fig:overview}
\end{figure*}

\subsection{Benchmarking Foundation Models}
Due to the increasing availability of large-scale pathology foundation models of varying size, there is an increasing demand for unified and objective benchmarks of such models. Benchmarking these models is essential for providing fair and transparent comparisons across different architectures, training strategies, and datasets. Recent efforts have focused on establishing standardized evaluation protocols and curated test sets that encompass a diverse range of clinically relevant tasks. 

\cite{Ma2025} introduced PathBench as a comprehensive benchmarking framework, featuring multi-center datsets, rigorous leakage prevention, and a standardized evaluation protocol across 64 diagnosis and prognosis tasks. They collected 15,888 slides from 8,549 patients and 10 hospitals and evaluated 19 recently published foundation models using a standardized preprocessing and linear probing protocol and showed that Virchow2 and H-optimus-1 are the most effective models overall. 

Similarly, \cite{Campanella2025} provide a clinical benchmark dataset collected during standard hospital operation from three health systems including tasks such as disease detection and biomarker prediction. They evaluated 11 foundation models concluding that all DINO and DINOv2 trained models perform comparably, where H-optimus-0 and Prov-GigaPath performed significantly better in a few tasks. 

\cite{Lee2025} performed a benchmark evaluation of four foundation models across 20 datasets in different scenarios where they address the influence of different adaptation strategies. In one scenario they tested linear probing, full fine-tuning, partial fine-tuning, \ac{peft} such as \ac{lora} \citep{Hu2022} and fully supervised learning and concluded that \ac{lora} was both most efficient and effective when adapting to diverse datasets within the same classification tasks. 

\cite{Breen2025} performed a task-specific benchmark study for ovarian cancer subtype classification. They compared three ImageNet-pretrained encoders and fourteen foundation models, each trained with 1,864 slides collected at Leeds Teaching Hospitals NHS Trust and validated on two external datasets. Their best performing classifier used the H-optimus-0, although UNI achieved similar results with only a quarter of the computational cost. 

\cite{Neidlinger2024} benchmarked 19 foundation models on 9,528 slides from lung, colorectal, gastric, and breast cancers. They evaluated 31 weakly-supervised tasks related to morphology, biomarkers and prognostication. They report that the vision-language model CONCH \citep{Lu2024} yielded the highest performance, when compared to vision-only models, where Virchow2 is the second best model. They also evaluated the downstream performance under different data scarcity settings. Their results indicate that while larger and more diverse pretraining datasets -- slide count, patient count, and tissue site diversity -- are generally associated with improved downstream performance, other factors like architecture and dataset quality also play critical roles. In scarce cohorts with only 75 to 300 patients for a specific downstream task or when rare biomarkers are involved, performance differences between models become more pronounced, with all models showing a decline as fine-tuning data decreases. 

\subsection{Benchmarking Adaptation Strategies}
Adapting large-scale foundation models with billions of parameters to specific downstream tasks remains a significant challenge, particularly in resource-constrained settings. Parameter-efficient fine-tuning techniques, such as \ac{lora} \citep{Hu2022}, have emerged as promising solutions to address these challenges by enabling effective adaptation with minimal additional parameters. \cite{Yang2024} provide a comprehensive review of \ac{lora}, discussing its applications and associated challenges. Despite encouraging results reported in studies such as \cite{Lee2025}, the use of \ac{lora} in medical applications remains underexplored. For instance, \cite{Cui2024} adapted a ViT-B model pretrained with DINOv2 using \ac{lora} for surgical depth estimation, demonstrating significant improvements over state-of-the-art models on the SCARED dataset which collected from da Vinci Xi endoscope surgery. Similarly, \cite{Dausort2024} investigated the application of \ac{lora} for cytological classification, evaluating five foundation models fine-tuned with \ac{lora} across four datasets. Their results show that \ac{lora} fine-tuning consistently outperforms linear probing, with particularly strong gains in few-shot scenarios where labeled data is scarce. Furthermore, in a dataset scaling experiment, they demonstrated that a CLIP model fine-tuned with \ac{lora} surpassed the state-of-the-art HierSwin \citep{Cai2024} model when trained on just 70\% of the data, highlighting the advantages of parameter-efficient adaptation methods like \ac{lora} when labeled data is limited or costly to acquire.

Despite these advances, there remains a critical need to systematically evaluate how foundation models adapt to clinically relevant tasks such as mitotic figure classification, particularly when labeled data is limited. The comparative effectiveness of linear probing and \ac{lora} for mitotic figure classification under varying data regimes remains an open question. 



\section{Methodology}

In this work, we address this gap by benchmarking recently published pathology foundation models on the task of mitotic figure classification. We focus on comparing linear probing and \ac{lora}-based fine-tuning in a data scaling experiment, highlighting their respective strengths and limitations in scarce data scenarios. Additionally, we investigate the robustness of these foundation models and adaptation strategies in a cross-domain setting, assessing their generalizability across different tumor types. 
 
\subsection{Datasets}
Recent publicly available mitotic figure datasets (e.g., CMC \citep{Aubreville2020}, \ac{ccmct} \citep{Bertram2019}, MIDOG 2022 \citep{Aubreville2024}, MIDOG++ \citep{Aubreville2023midogpp}) are larger and more diverse than before, making them ideal for benchmarking adaptation strategies. We conduct our analysis on two of these datasets, \ac{ccmct} \citep{Bertram2019} and MIDOG 2022 \citep{Aubreville2024} (Table \ref{tab:datasets}), each chosen for their complementary properties. The \ac{ccmct} dataset is a large-scale resource focused on a single tumor domain, making it particularly well-suited for scaling experiments and in-depth analysis within a consistent biological context. In contrast, MIDOG 2022 is a highly diverse dataset spanning multiple tumor types, species, and laboratories, providing an ideal benchmark for evaluating model robustness and generalization in cross-domain settings.
 
 \subsubsection{CCMCT}
 The \ac{ccmct} dataset consists out of 32 fully annotated \ac{ccmct} \acp{wsi} with $44,880$ annotations for mitotic figures and $27,965$ hard negatives, including both low grade cases as well as high grade cases. The images were scanned with an Aperio ScanScope S2 \ac{wsi} scanner at a resolution of $0.25$ microns per pixel. For some examples see Figure \ref{fig:overview}.

 \subsubsection{MIDOG 2022}
The MIDOG 2022 dataset is a comprehensive multi-tumor, multi-species, and multi-laboratory collection comprising 354 \acp{roi} with a total of $11,051$ mitotic figures and $9,501$ challenging negative samples. It includes five distinct tumor types: 
canine cutaneous mast cell tumor (domain A), scanned at 40x magnification ($0.25$ microns per pixel) using the Aperio ScanScope CS2; canine lymphoma (domain B) scanned with the 3DHistech Panoramic Scan II scanner at $0.25$ microns per pixel; human breast cancer (domain C), with 50 slides each scanned using Hamamatsu XR, Hamamatsu S360, and Aperio ScanScope CS2 scanners at resolutions ranging from $0.23$ to $0.25$ microns per pixel;
human neuroendocrine tumor (domain D), scanned at 40x ($0.23$ microns per pixel) with the Hamamatsu XR; and canine lung cancer (domain E) digitized with the 3DHistech Panoramic Scan II scanner at $0.25$ microns per pixel. Some examples are shown in Figure \ref{fig:overview}.

\begin{table}[ht]
\caption{Summary of datasets.}
\label{tab:datasets}
\resizebox{\columnwidth}{!}{%
\begin{tabular}{@{}lllllll@{}}
\toprule
Dataset    & Images & Mitotic figures & Hard negatives & Tumor types & Scanner & Magnification \\ \midrule
CCMCT      & 32 WSIs    & 44,880          & 27,965         & 1           & 1       & 40x           \\
MIDOG 2022 & 354 ROIs   & 11,051          & 9,501          & 5           & 4       & 40x           \\ \bottomrule
\end{tabular}%
}
\end{table}

\subsection{Foundation Models}

We selected six state-of-the-art pathology foundation models (Table \ref{tab:models}) that represent a diverse range of architectures, pretraining strategies, and dataset scales. The models include Phikon \citep{Filiot2023}, UNI \citep{Chen2024UNI}, Virchow \citep{Vorontsov2023}, Virchow2 \citep{Zimmermann2024}, H-optimus-0 \citep{Saillard2024}, and Prov-GigaPath \citep{Xu2024}, spanning backbones from ViT-B to ViT-G and pretraining algorithms such as iBOT and DINOv2. These models were pretrained on datasets ranging from public sources like TCGA to large proprietary collections, with slide counts varying from 6,093 to over 3 million and tile counts from 43 million to 2 billion. The selection covers a broad spectrum of model sizes (86M to 1B parameters), pretraining magnifications, and feature dimensionalities, providing a comprehensive benchmark for mitotic figure classification across different data and model scales. Additionally, we compare the pathology foundation models with ViT-S DINOv3 \citep{siméoni2025dinov3}, a compact self-supervised Vision Transformer from the DINOv3 family. Trained on large-scale web datasets, it serves as a general-purpose visual encoder that produces robust, high-quality dense representations transferable across diverse vision tasks without task-specific fine-tuning.

\begin{table*}[h]
\caption{Summary of pathology foundation models.}
\resizebox{\textwidth}{!}{%
\begin{tabular}{@{}lllllllll@{}}
\toprule
Model         & Backbone & Pretraining algorithm & Parameters & Data source & Pretraining magnifications & Slide count & Tile count & Patch features \\ \midrule
Phikon        & ViT-B    & iBOT                  & 86M        & TCGA        & 20x                        & 6093        & 43,3M      & 768            \\
UNI           & ViT-L    & DINOv2                & 304M       & Proprietary & 20x                        & 100K        & 100M       & 1024           \\
Virchow       & ViT-H    & DINOv2                & 632M       & Proprietary & 20x                        & 1.5M        & 2B         & 1280           \\
Virchow2      & ViT-H    & DINOv2                & 632M       & Proprietary & 5x, 10x, 20x, 40x          & 3.1M        & 1.7B       & 1280           \\
H-optimus-0   & ViT-G    & DINOv2                & 1B         & Proprietary & 20x                        & 500K        & NA         & 1536           \\
Prov-GigaPath & ViT-G    & DINOv2               & 1B         & Proprietary & 20x                        & 170K        & 1.3B       & 1536           \\ \bottomrule
\end{tabular}%
}
\label{tab:models}
\end{table*}

\subsection{Linear Probing}
Linear probing is a widely adopted and straightforward approach for evaluating the quality of representations learned by pre-trained models. In this method, the parameters of a pre-trained model $\theta$ are frozen, and only a linear classifier is trained on top of the extracted features to adapt the model to a specific downstream task. Given an input $x$, the model computes feature representations $z = f_\theta(x)$, which are then passed to a linear layer. The output of the classifier is given by:

\begin{equation}
    y = \mathbf{W}z
\end{equation}

where $z\in \mathbb{R}^n$ is the feature vector, and $\mathbf{W} \in \mathbb{R}^{c \times n}$ the weight matrix of the linear classifier,  with $n$ the number of extracted features and $c$ the number of classes. Linear probing serves as a standard benchmark to assess how well the pre-trained features can be separated by a simple linear decision boundary, providing insight into the generalizability and utility of the learned representations for new tasks without updating the backbone model. This approach is particularly valuable in scenarios with limited labeled data, as it requires training only a small number of parameters.

\subsection{Low-Rank Adaptation}
To explore the benefits of fine-tuning foundation models beyond the classification head, we employ Low-Rank Adaptation (LoRA) \citep{Hu2022}, a parameter-efficient fine-tuning technique. \Ac{lora} updates pre-trained model weights using low-rank decomposition. Instead of directly modifying the full weight matrix $\mathbf{W} \in \mathbb{R}^{m \times n}$, \ac{lora} introduces a weight correction $\Delta \mathbf{W} \in \mathbb{R}^{m \times n}$, which is expressed as the product of two smaller matrices: $\mathbf{A} \in \mathbb{R}^{r \times n}$ and $\mathbf{B} \in \mathbb{R}^{m \times r}$, where $r$ is the chosen rank and typically much smaller than $m$ or $n$. The forward pass is then modified as follows:

\[
    h = \mathbf{W}x + \gamma \Delta \mathbf{W}x = \mathbf{W}x + \gamma \mathbf{B}\mathbf{A}x
\]

Here, $h$ denotes the hidden state at a given layer, $x$ is the input to that layer, and $\gamma$ is a scaling factor. During initialization, $\mathbf{A}$ is randomly initialized, while $\mathbf{B}$ is set to zero. Only the entries of these low-rank matrices are updated during training, while the original model weights remain frozen, significantly reducing the number of trainable parameters. At inference time, the effective weight matrix is simply the sum of the original and the low-rank update, so the computational cost remains unchanged.

\subsection{Baselines}
We compare the performance of the foundation models to multiple baselines. First, we use the two widely adopted feature extractors ResNet50 and ViT-H, pretrained on ImageNet, to generate embeddings for linear probing in the same manner as with the foundation models. Additionally, we train four models starting from ImageNet pretraining in a fully supervised setting. We select a ResNet50 and three Vision Transformer (ViT-S, ViT-B, ViT-H) covering a range of model scales to directly compare with their foundation model counterparts. For these supervised baselines, standard image augmentations are applied during training, including random color jitter, Gaussian blur, flipping, and random rotations. This setup allows for a direct comparison between foundation model adaptation strategies and traditional supervised learning approaches.

\subsection{Dataset Scaling Experiment}

To systematically evaluate how foundation models and adaptation strategies perform under varying data availability, we conducted a dataset scaling experiment using both the large-scale, single-domain \ac{ccmct} dataset and the diverse, multi-domain MIDOG 2022 dataset. For each model, we trained on four different fractions of the available data (0.1\%, 1\%, 10\%, and 100\%), enabling us to assess model robustness and adaptation in both data-rich and data-scarce scenarios. To enhance the statistical reliability of our findings, we employed five-fold Monte Carlo cross-validation for each combination of model and training set size, resulting in 20 independent training runs per model. In the beginning, 20\% of all annotations from the respective dataset were randomly selected as the test set to ensure a fair comparison between the different fractions of the data. Then, for each run, we set aside 20\% of the remaining training annotations for validation. We used a case-level split to avoid data leakage between the splits. All models were trained and evaluated on identical splits to ensure fair comparison between the models. It is important to note that, due to the smaller overall size of MIDOG 2022, the absolute number of training samples at each percentage level was substantially lower than in \ac{ccmct}, providing a stringent test of model performance in low-data regimes.

\subsection{Cross-Domain Experiment}

To further investigate the generalization capabilities of the models, we performed a cross-domain experiment using the MIDOG 2022 dataset, which is especially suitable for multi-domain evaluation. In this setup, we used each of the five tumor domains as the training domain, while using the remaining domains exclusively for testing, thereby simulating real-world scenarios where models are deployed on previously unseen data distributions. For in-domain evaluation, 20\% of the data from the training domain was withheld and included as further test set for in-domain evaluation. As with the scaling experiment, we conducted five independent training runs per domain and per model, resulting in a total of 25 training sessions per model. This experimental design allows for a comprehensive assessment of both in-domain and cross-domain robustness, leveraging the diversity of MIDOG 2022.

\begin{figure*}[!ht]
    \centering
    \includegraphics[width=\textwidth]{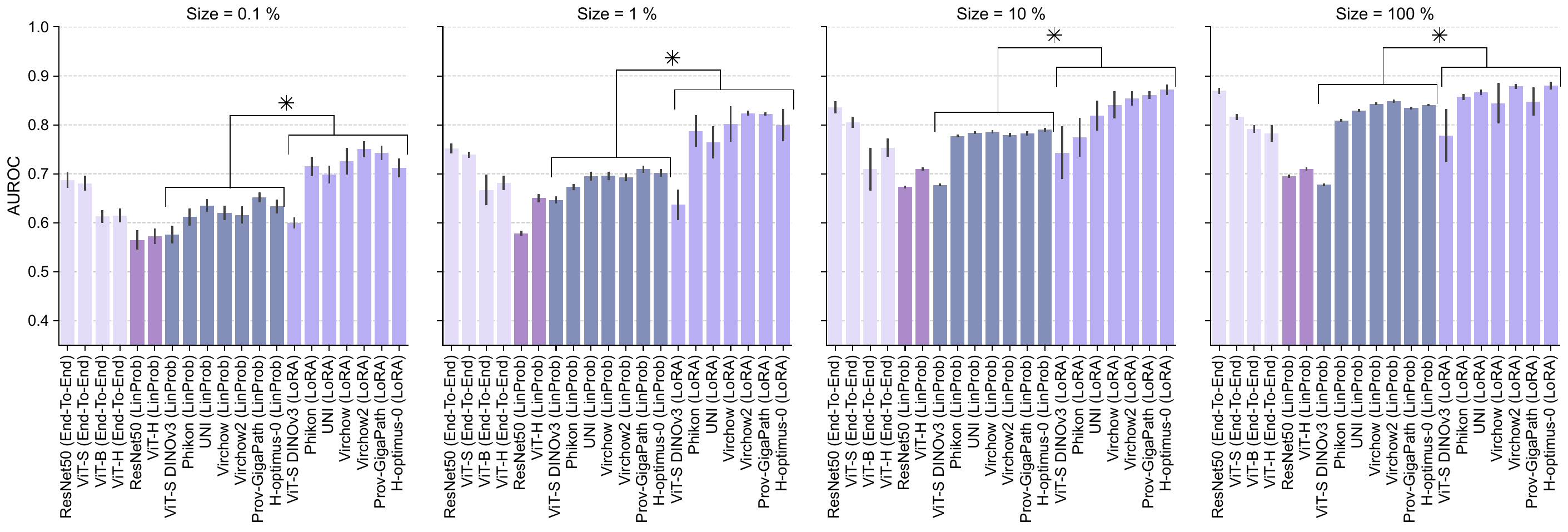}
    \caption{Results of the data scaling experiment on the CCMCT dataset. (*) indicates statistical significance ($\alpha < 0.05$) between the pooled scores of LoRA and LinProb models.}
    \label{fig:CCMCT_datascale}
\end{figure*}

\begin{figure*}[!ht]
    \centering
    \includegraphics[width=\textwidth]{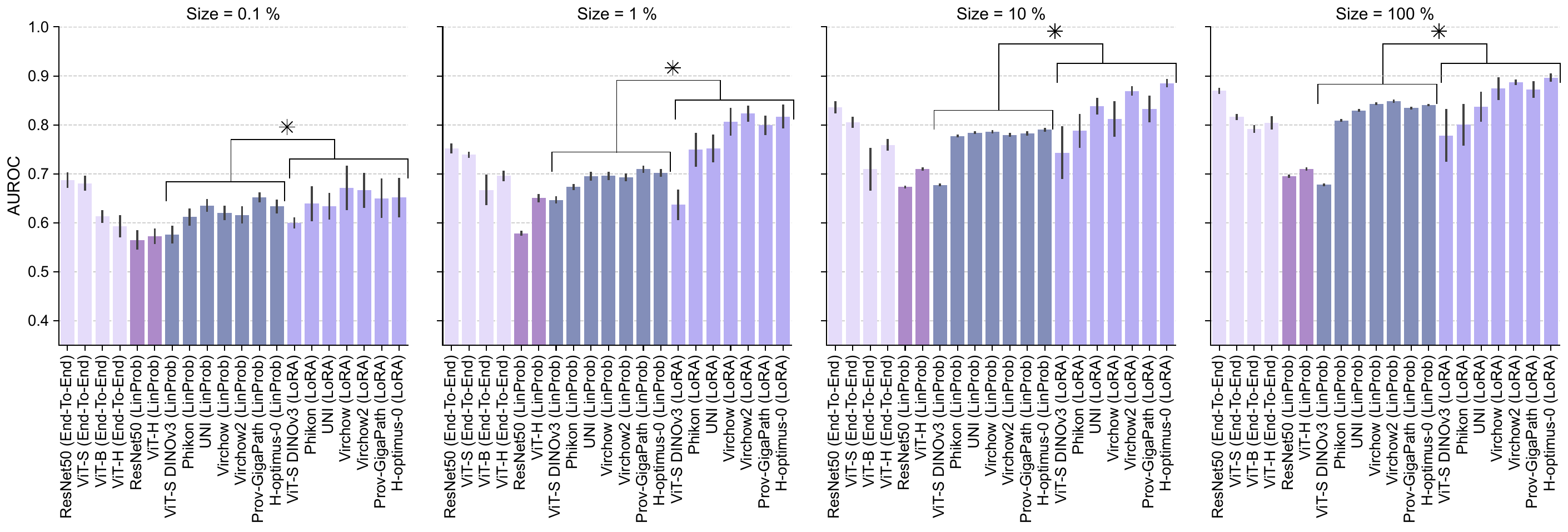}
    \caption{Results of the data scaling experiment on the MIDOG 2022 dataset. (*) indicates statistical significance ($\alpha < 0.05$) between the pooled scores of LoRA and LinProb models.}
    \label{fig:MIDOG_datascale}
\end{figure*}

\subsection{Implementation Details}
For the mitotic figure classification task we define $c=2$ to distinguish between mitotic figures and mitotic figure look-alikes (hard negatives), which both datasets provide. We extracted image patches of size $224 \times 224$ pixels centered around the mitotic figure and hard negative annotations \citep{Vorontsov2023}. The patch embeddings for the linear probing experiments were created by passing the patches through the frozen encoder of each model. The patches were normalized according to the means and standard deviations provided in the respective works. No additional data augmentation was applied. For Virchow and Virchow2, these patch embeddings were created by concatenating the class token and the mean across all other 256 predicted patch tokens, as described in \cite{Vorontsov2023}. For all other models, only the class token was used \citep{Vorontsov2023}. We use the linear probing implementation by \cite{Chen2024UNI}.

We apply \ac{lora} to the query, key, and value projection layer and the output projection layer of the attention blocks and the first and second fully connected layers in the multi-layer perceptron (MLP) sections. The rank is set to 16, the scaling factor is set to 16, the dropout is set to $0.1$.

The fully supervised learning of the baselines and the \ac{lora}-adaptation of the foundation models is performed using the Adam optimizer with default parameter values ($\beta_1=0.9$, $\beta_2=0.999$, $\epsilon=1e-8$), batch size of 16, patch size of $224\times 224$, and standard image augmentations. Binary cross-entropy loss is adopted as loss function. In each pseudo epoch, patches are sampled randomly with 50\% of the patches containing mitotic figures, and the other 50\% containing either a hard negative patch or a completely random patch, each with 25\% probability. The pseudo-epoch length is set to 1280. We train the models for 100 pseudo epochs using a one-cycle learning rate policy \citep{smith2017super}, including a linear warum-up phase followed by cosine annealing, with a maximum learning rate of $10^{-4}$, and select the best model retrospectively based on the validation loss per epoch. During inference, we evaluate only patches that contain annotations for mitotic figures or hard negatives. The classification experiments are evaluated with balanced accuracy and weighted F1 scores due to the class imbalance between mitotic figures and hard negatives, and additionally with the AUROC score. All experiments were executed on a workstation equipped with a single NVIDIA RTX 3090 GPU.

\section{Results}

\subsection{Data Scaling Experiment}

The results of the data scaling experiment are shown in Figure \ref{fig:CCMCT_datascale} and Figure \ref{fig:MIDOG_datascale}. The experiments on both the single-domain \ac{ccmct} and the multi-domain MIDOG 2022 datasets consistently demonstrate the superior adaptability and data efficiency of pathology foundation models, particularly when adapted with parameter efficient fine-tuning methods such as \ac{lora}. Across both datasets, foundation models outperform traditional feature extractors such as ResNet50 and ViT-H pretrained on ImageNet at every data regime, with the performance gap being most pronounced in low-data settings. 

On the \ac{ccmct} dataset, which represents a large-scale, single-domain scenario, foundation models fine-tuned with \ac{lora} achieve substantial gains in AUROC as the training set size increases. Already at the smallest data fraction of 0.1\% the \ac{lora} adapted foundation models surpass the baselines, and this advantage becomes more pronounced as more data is made available. At full data scale, these models approach AUROC values of 0.9, while baselines plateau at lower levels. Interestingly, the performance of \ac{lora} fine-tuned foundation models already reaches near full-data scale performance already at 10\% of the available data, especially with the H-optimus-0 model. The ResNet50 End-to-End model consistently outperforms all foundation models adapted with standard linear probing and is only outperformed by Virchow2 and H-optimus-0 adapted with \ac{lora} at full-data scale (Table \ref{tab:CCMCT_full_datascale}). The ViT End-to-End variants show similar performance across the different data sizes with a clear advantage of the more compact ViT-S model at all data scales. However, their AUROC scores are considerably lower compared to the ResNet50 End-to-End model and moderately lower compared to their foundation model counterparts, especially for ViT-B and ViT-H. The performance of adapted ViT-S DINOv3 model lacks behind the pathology foundation models in both linear probing and \ac{lora} settings. Even in the \ac{lora} setting, the performance cannot match its ViT-S End-to-End counterpart, indicating that the domain shift from general purpose weights to histopathology data requires more extensive adaptation than \ac{lora} fine-tuning. 

A similar trend is observed on the MIDOG 2022 dataset, which is characterized by greater diversity in tumor types, species, and imaging conditions. Compared to \ac{ccmct}, the more challenging conditions and reduced absolute training sample sizes in MIDOG 2022 are reflected in generally lower AUROC scores, particularly for linear probing baselines and foundation models at the lowest data regimes (0.1\% to 1\%). However, the benefits of foundation models with \ac{lora} adaptation become more clear at higher data fractions (1\% to 100\%), where these models substantially outperform their linear probing counterparts. The only exceptions are UNI and Phikon, which lag slightly behind at full data scale. Again, the ResNet50 End-to-End baseline outperforms linear probing of all foundation models and is only outperformed by Virchow2 and H-optimus-0 adapted with \ac{lora} at full data scale (Table \ref{tab:MIDOG_full_datascale}). 

For each data scale, we pooled the performance scores from all foundation models fine‑tuned using linear probing or \ac{lora}, yielding 35 paired observations per adaptation strategy (7 models $\times$ 5 repetitions). To assess whether performance differed between the two adaptation strategies, we conducted Wilcoxon signed‑rank tests \cite{Wilcoxon1945}. Across all data scales, the results indicated that linear probing achieved significantly lower scores than \ac{lora} ($\alpha = 0.05$).

\begin{table}[!ht]
    \centering
    \caption{Results at 100 \% dataset size of CCMCT dataset.}
    \label{tab:CCMCT_full_datascale}
    \resizebox{\columnwidth}{!}{%
      \begin{tabular}{lccc}
        \toprule
        Model & AUROC & Balanced ACC & Weighted F1 \\
        \midrule
        ResNet50 (End-To-End) & 0.87$\pm$0.01 & 0.79$\pm$0.01 & 0.78$\pm$0.01 \\
        ViT-S (End-To-End) & 0.82$\pm$0.01 & 0.73$\pm$0.01 & 0.72$\pm$0.03 \\
        ViT-B (End-To-End) & 0.79$\pm$0.01 & 0.70$\pm$0.02 & 0.71$\pm$0.03 \\
        ViT-H (End-To-End) & 0.78$\pm$0.03 & 0.67$\pm$0.04 & 0.67$\pm$0.11 \\ \midrule
        ResNet50 (LinProb) & 0.69$\pm$0.00 & 0.61$\pm$0.00 & 0.65$\pm$0.00 \\
        ViT-H (LinProb) & 0.71$\pm$0.00 & 0.61$\pm$0.00 & 0.65$\pm$0.00 \\ \midrule
        ViT-S DINOv3 (LinProb) & 0.68$\pm$0.00 & 0.57$\pm$0.00 & 0.60$\pm$0.00 \\
        Phikon (LinProb) & 0.81$\pm$0.00 & 0.71$\pm$0.00 & 0.74$\pm$0.00 \\
        UNI (LinProb) & 0.83$\pm$0.00 & 0.73$\pm$0.00 & 0.76$\pm$0.00 \\
        Virchow (LinProb) & 0.84$\pm$0.00 & 0.75$\pm$0.00 & 0.78$\pm$0.00 \\
        Virchow2 (LinProb) & 0.85$\pm$0.00 & 0.76$\pm$0.00 & 0.78$\pm$0.00 \\
        Prov-GigaPath (LinProb) & 0.83$\pm$0.00 & 0.74$\pm$0.00 & 0.77$\pm$0.00 \\
        H-optimus-0 (LinProb) & 0.84$\pm$0.00 & 0.75$\pm$0.00 & 0.78$\pm$0.00 \\ \midrule
        ViT-S DINOv3 (LoRA) & 0.78$\pm$0.12 & 0.70$\pm$0.09 & 0.71$\pm$0.09 \\
        Phikon (LoRA) & 0.86$\pm$0.01 & 0.77$\pm$0.01 & 0.78$\pm$0.01 \\
        UNI (LoRA) & 0.87$\pm$0.01 & 0.78$\pm$0.01 & 0.78$\pm$0.01 \\
        Virchow (LoRA) & 0.84$\pm$0.09 & 0.75$\pm$0.07 & 0.76$\pm$0.06 \\
        Virchow2 (LoRA) & 0.88$\pm$0.01 & 0.78$\pm$0.01 & 0.80$\pm$0.01 \\
        Prov-GigaPath (LoRA) & 0.85$\pm$0.06 & 0.76$\pm$0.05 & 0.77$\pm$0.04 \\
        H-optimus-0 (LoRA) & \textbf{0.88$\pm$0.01} & \textbf{0.79$\pm$0.02} & \textbf{0.80$\pm$0.01} \\
        \bottomrule
    \end{tabular}%
    }
\end{table}

\begin{table}[!ht]
    \centering
    \caption{Results at 100\% dataset size of MIDOG dataset.}
    \label{tab:MIDOG_full_datascale}
    \resizebox{\columnwidth}{!}{%
         \begin{tabular}{lccc}
        \toprule
        Model & AUROC & Balanced ACC & Weighted F1 \\
        \midrule
        ResNet50 (End-To-End) & 0.87$\pm$0.009 & 0.79$\pm$0.009 & 0.78$\pm$0.010 \\
        ViT-S (End-To-End) & 0.82$\pm$0.010 & 0.73$\pm$0.012 & 0.72$\pm$0.027 \\
        ViT-B (End-To-End) & 0.79$\pm$0.013 & 0.70$\pm$0.021 & 0.71$\pm$0.025 \\
        ViT-H (End-To-End) & 0.80$\pm$0.035 & 0.70$\pm$0.051 & 0.70$\pm$0.087 \\ \midrule
        ResNet50 (LinProb) & 0.69$\pm$0.003 & 0.61$\pm$0.002 & 0.65$\pm$0.002 \\
        ViT-H (LinProb) & 0.71$\pm$0.003 & 0.61$\pm$0.002 & 0.65$\pm$0.002 \\  \midrule
        ViT-S DINOv3 (LinProb) & 0.68$\pm$0.002 & 0.57$\pm$0.003 & 0.60$\pm$0.005 \\
        Phikon (LinProb) & 0.81$\pm$0.001 & 0.71$\pm$0.002 & 0.74$\pm$0.002 \\
        UNI (LinProb) & 0.83$\pm$0.001 & 0.73$\pm$0.003 & 0.76$\pm$0.002 \\
        Virchow (LinProb) & 0.84$\pm$0.001 & 0.75$\pm$0.002 & 0.78$\pm$0.002 \\
        Virchow2 (LinProb) & 0.85$\pm$0.002 & 0.76$\pm$0.001 & 0.78$\pm$0.001 \\
        Prov-GigaPath (LinProb) & 0.83$\pm$0.001 & 0.74$\pm$0.002 & 0.77$\pm$0.002 \\
        H-optimus-0 (LinProb) & 0.84$\pm$0.001 & 0.75$\pm$0.002 & 0.78$\pm$0.002 \\ \midrule
        ViT-S DINOv3 (LoRA) & 0.78$\pm$0.115 & 0.70$\pm$0.092 & 0.71$\pm$0.095 \\
        Phikon (LoRA) & 0.80$\pm$0.128 & 0.73$\pm$0.102 & 0.73$\pm$0.105 \\
        UNI (LoRA) & 0.84$\pm$0.090 & 0.76$\pm$0.072 & 0.76$\pm$0.072 \\
        Virchow (LoRA) & 0.87$\pm$0.067 & 0.78$\pm$0.060 & 0.79$\pm$0.051 \\
        Virchow2 (LoRA) & 0.89$\pm$0.011 & 0.80$\pm$0.022 & 0.81$\pm$0.014 \\
        Prov-GigaPath (LoRA) & 0.87$\pm$0.047 & 0.79$\pm$0.044 & 0.79$\pm$0.037 \\
        H-optimus-0 (LoRA) & \textbf{0.90$\pm$0.019} & \textbf{0.81$\pm$0.022} & \textbf{0.81$\pm$0.015} \\
        \bottomrule
        \end{tabular}%
        }
\end{table}

\subsection{Cross-Domain Experiment}

\begin{table*}[!ht]
\caption{Results of the cross-domain experiment. The results are averaged across all in-domain and out-domain scenarios. Displayed are the mean score and the standard deviation. ($^{\ast}$) indicates statistically significant differences compared to the LinProb counterpart ($\alpha=0.05$). ($^{\dagger}$) indicates statistically significant differences compared to the ResNet50 End-to-End baseline ($\alpha=0.05$).}
\label{tab:cross_domain}
\resizebox{\textwidth}{!}{%
\begin{tabular}{@{}lllllll@{}}
\toprule
\multirow{2}{*}{Model}  & \multicolumn{2}{l}{AUROC}       & \multicolumn{2}{l}{Balanced ACC} & \multicolumn{2}{l}{Weighted F1} \\ \cmidrule(l){2-7} 
                        & In-domain     & Out-of-domain   & In-domain      & Out-of-domain   & In-domain      & Out-of-domain  \\ \midrule
ResNet50 (End-to-End)   & 0.87$\pm$0.04 & 0.74 $\pm$0.07  & 0.79$\pm$0.03  & 0.67$\pm$0.06   & 0.81$\pm$0.03  & 0.67$\pm$0.09  \\
ViT-S (End-to-End)      & 0.84$\pm$0.03 & 0.76$\pm$0.04   & 0.74$\pm$0.03  & 0.68$\pm$0.04   & 0.75$\pm$0.02  & 0.67$\pm$0.06  \\
ViT-B (End-to-End)      & 0.83$\pm$0.03 & 0.75$\pm$0.05   & 0.75$\pm$0.03  & 0.67$\pm$0.04   & 0.76$\pm$0.04  & 0.65$\pm$0.07  \\ 
ViT-H (End-to-End)      & 0.83$\pm$0.03 & 0.75$\pm$0.04   & 0.75$\pm$0.02  & 0.66$\pm$0.04   & 0.76$\pm$0.03  & 0.65$\pm$0.08  \\ \midrule
ResNet50 (LinProb)     & 0.63$\pm$0.04 & 0.53$\pm$0.03   & 0.58$\pm$0.02  & 0.51$\pm$0.01   & 0.62$\pm$0.05  & 0.48$\pm$0.08  \\
ViT-H (LinProb)         & 0.68$\pm$0.05 & 0.56$\pm$0.04   & 0.59$\pm$0.03  & 0.53$\pm$0.02   & 0.63$\pm$0.05  & 0.49$\pm$0.10  \\ \midrule
ViT-S DINOv3 (LinProb)  & 0.65$\pm$0.06 & 0.58$\pm$0.04   & 0.56$\pm$0.02  & 0.53$\pm$0.03   & 0.58$\pm$0.07  & 0.47$\pm$0.12  \\
Phikon (LinProb)        & 0.76$\pm$0.03 & 0.61$\pm$0.05   & 0.68$\pm$0.03  & 0.56$\pm$0.04   & 0.71$\pm$0.03  & 0.55$\pm$0.09  \\
UNI (LinProb)           & 0.76$\pm$0.03 & 0.64$\pm$0.05   & 0.69$\pm$0.03  & 0.59$\pm$0.03   & 0.71$\pm$0.02  & 0.59$\pm$0.06  \\
Virchow (LinProb)       & 0.79$\pm$0.04 & 0.66$\pm$0.06   & 0.72$\pm$0.04  & 0.60$\pm$0.03   & 0.74$\pm$0.02  & 0.60$\pm$0.06  \\
Virchow2 (LinProb)      & 0.79$\pm$0.04 & 0.66$\pm$0.05   & 0.71$\pm$0.04  & 0.60$\pm$0.04   & 0.74$\pm$0.02  & 0.59$\pm$0.07  \\
Prov-GigaPath (LinProb) & 0.79$\pm$0.03 & 0.66$\pm$0.07   & 0.71$\pm$0.03  & 0.61$\pm$0.04   & 0.74$\pm$0.02  & 0.61$\pm$0.05  \\
H-optimus-0 (LinProb)   & 0.79$\pm$0.04 & 0.66$\pm$0.07   & 0.72$\pm$0.03  & 0.60$\pm$0.05   & 0.74$\pm$0.02  & 0.61$\pm$0.07  \\ \midrule
ViT-S DINOv3 (LoRA)     & 0.82$\pm$0.05$^{\ast}$         & 0.75$\pm$0.05$^{\ast}$           & 0.74$\pm$0.04$^{\ast \dagger}$  & 0.68$\pm$0.04$^{\ast \dagger}$   & 0.74$\pm$0.04$^{\ast \dagger}$  & 0.67$\pm$0.06$^{\ast \dagger}$  \\
Phikon (LoRA)           & 0.81$\pm$0.05$^{\ast}$         & 0.76$\pm$0.09$^{\ast}$           & 0.74$\pm$0.07                   & 0.69$\pm$0.07                    & 0.76$\pm$0.07                   & 0.68$\pm$0.09  \\
UNI (LoRA)              & 0.86$\pm$0.05$^{\ast \dagger}$ & 0.80$\pm$0.06$^{\ast \dagger}$   & 0.78$\pm$0.05                   & 0.72$\pm$0.06                    & 0.79$\pm$0.04$^{\ast}$   	   & 0.72$\pm$0.08$^{\ast}$  \\
Virchow (LoRA)          & 0.86$\pm$0.08$^{\ast \dagger}$ & 0.82$\pm$0.08$^{\ast \dagger}$   & 0.78$\pm$0.06                   & 0.75$\pm$0.07                    & 0.79$\pm$0.07                   & 0.75$\pm$0.09  \\
Virchow2 (LoRA)         & 0.89$\pm$0.02$^{\ast \dagger}$ & 0.87$\pm$0.02$^{\ast \dagger}$   & 0.82$\pm$0.02$^{\ast}$          & 0.79$\pm$0.03$^{\ast}$           & 0.83$\pm$0.01$^{\ast \dagger}$  & 0.80$\pm$0.05$^{\ast \dagger}$  \\
Prov-GigaPath (LoRA)    & 0.89$\pm$0.02$^{\ast \dagger}$ & 0.85$\pm$0.04$^{\ast \dagger}$   & 0.80$\pm$0.03$^{\ast}$          & 0.76$\pm$0.05$^{\ast}$           & 0.81$\pm$0.02$^{\ast}$          & 0.77$\pm$0.06$^{\ast}$  \\
H-optimus-0 (LoRA)      & \bf{0.90$\pm$0.02$^{\ast \dagger}$} & \bf{0.88$\pm$0.02$^{\ast \dagger}$}   & \bf{0.82$\pm$0.02$^{\ast}$}  & \bf{0.80$\pm$0.03$^{\ast}$}   & \bf{0.83$\pm$0.01$^{\ast \dagger}$}  & \bf{0.81$\pm$0.04$^{\ast \dagger}$}  \\ \bottomrule
\end{tabular}%
}
\end{table*}

\begin{figure*}[!ht]
    \centering
    \includegraphics[width=\textwidth]{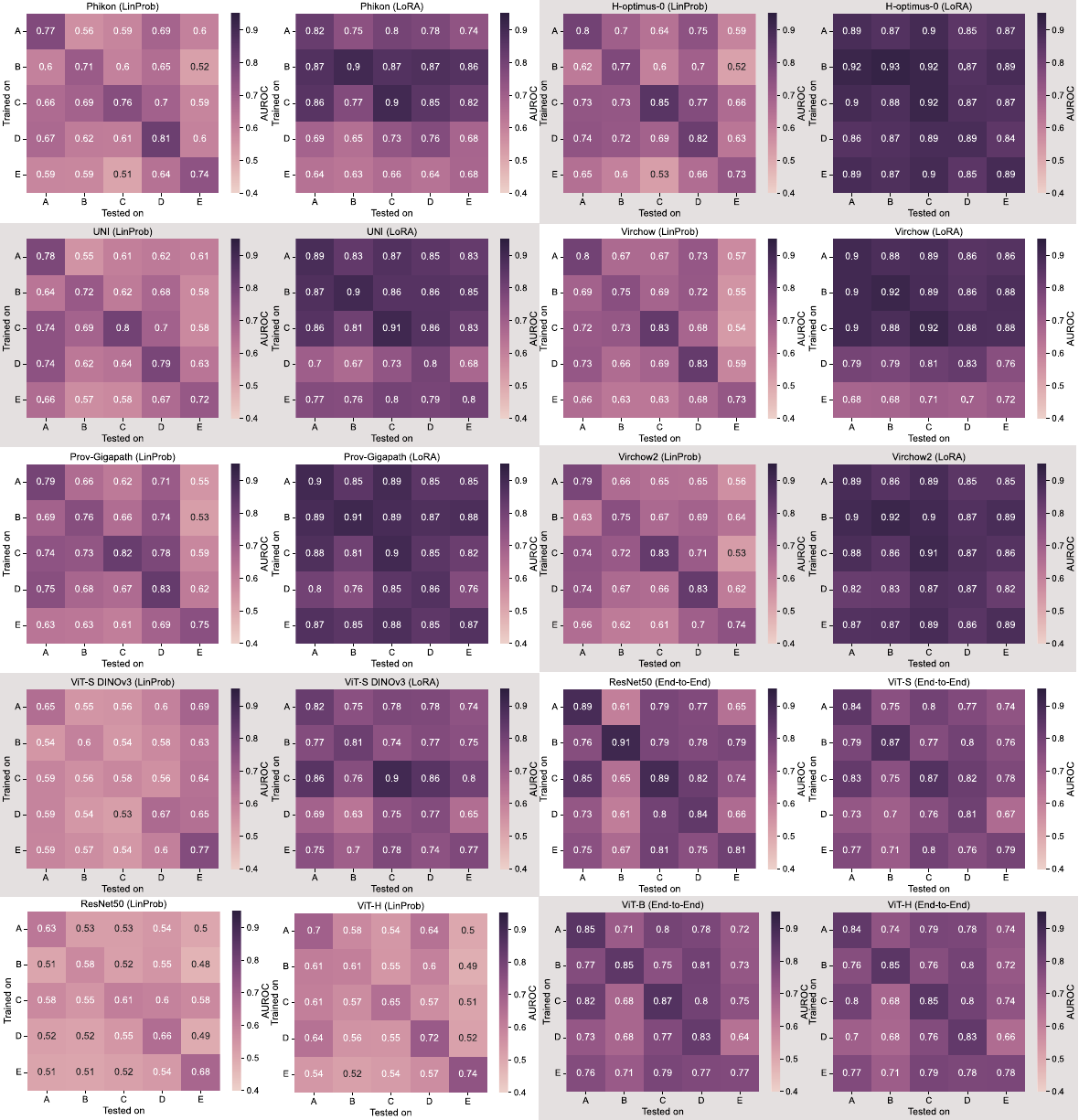}
    \caption{Results of the cross-domain experiment. We show each individual scenario with its averaged AUROC score over all training sessions. A: canine mast cell tumor. B: canine lymphoma. C: human breast cancer. D: human neuroendocrine tumor. E: canine lung cancer.}
    \label{fig:cross_domain}
\end{figure*}

The results of the cross-domain experiment are summarized in Table \ref{tab:cross_domain} and Figure \ref{fig:cross_domain}. As expected, all models perform better in-domain than out-of-domain, reflecting the inherent challenge of domain shifts in histopathology. Traditional feature extractors such as ResNet50 and ViT-H pretrained on ImageNet show limited generalization, with out-of-domain AUROC scores dropping to 0.53 and 0.56, respectively.

Foundation models adapted with standard linear probing demonstrate moderate improvements, with out-of-domain AUROC values in the range of 0.61-0.66 (Table \ref{tab:cross_domain}). However, the most substantial gains are observed when foundation models are fine-tuned with \ac{lora}. In this setting, models such as H-optimus-0, Virchow2, and Prov-Gigapath achieve out-of-domain AUROC scores of 0.88, 0.87, and 0.85 respectively, with only a minimal drop compared to their in-domain performance. This trend is consistent across all evaluation metrics, including balanced accuracy and weighted F1, highlighting the effectiveness of \ac{lora} in enhancing cross-domain robustness. 

We performed planned pairwise comparisons to assess whether the performance differences between the linear probing and \ac{lora} models were statistically significant, and whether \ac{lora} performance differed from the ResNet50 End‑to‑End baseline, yielding 14 model comparisons. For each in‑domain comparison, we analyzed 25 paired observations (5 domains × 5 repetitions), and for each out‑of‑domain comparison, 100 paired observations (5 domains × 4 out‑of‑domain evaluations × 5 repetitions). All planned comparisons were conducted separately for each evaluation metric and domain setting (in‑domain and out‑of‑domain) using the Wilcoxon signed‑rank test. Resulting p‑values were adjusted for multiple comparisons using the Holm procedure \citep{Holm1979} for each metric and domain setting. Nearly all \ac{lora} models significantly outperformed their linear‑probing counterparts for every evaluation metric, except for Phikon, UNI, and Virchow, where not all differences were statistically significant. The general‑purpose ViT‑S DINOv3 model was also significantly outperformed by the ResNet50 End‑to‑End baseline. The strongest improvements over the ResNet50 End‑to‑End baseline were observed for Virchow2 and H‑optimus‑0, where both F1‑scores and AUROC metrics differed significantly.

Looking at individual scenarios in Figure \ref{fig:cross_domain} we can clearly observe the strong gains of \ac{lora} adaptation for the foundation models. Especially when looking at models such as Prov-Gigapath, H-opimus-0, and Virchow2 where the \ac{lora}-adapted models nearly closed the out-of-domain performance gap between any scenario.  Despite strong gains with \ac{lora} some scenarios were still particularly challenging. Domain D (human neuroendocrine tumor) and E (canine lung cancer) were most challenging for all models. Training on either of these domains led to the worst performances across all models, with the biggest differences observed in Phikon, UNI, and Virchow.

\section{Discussion}

Our systematic benchmarking of pathology foundation models for mitotic figure classification across both single-domain and multi-domain datasets provides several important insights for the field of computational pathology. Most notably, our results demonstrate that foundation models, particularly when adapted with parameter-efficient fine-tuning methods such as \ac{lora}, offer substantial advantages in both data-scarce and cross-domain scenarios.

The data scaling experiments reveal that foundation models consistently outperform traditional feature extractors, with the performance gap being most pronounced in low-data regimes. This suggests that rich, transferable representations learned during large-scale pretraining can be effectively leveraged for a new task such as mitotic figure classification when only a small amount of labeled data is available. The ability of \ac{lora}-adapted models to reach near full-data scale performance with as little as 10\% of the available data highlights the practical value of parameter-efficient adaptation strategies for real-world applications where annotation is often costly and time-consuming. Furthermore, the comparison between fully fine-tuning large vision transformers such as ViT-B and ViT-H and their \ac{lora}-adapted counterparts highlights how efficient \ac{lora} acts as a regularization when fine-tuning such large-scale models to a new task. On the other hand, the more commonly used linear probing does not utilize the full potential of foundation models, demonstrating inferior performance to \ac{lora}-adapted models and fully fine-tuned traditional architectures such as ResNet50. 

In the more challenging, heterogeneous setting of the MIDOG 2022 dataset, foundation models again demonstrate superior robustness, with \ac{lora} adaptation further narrowing the gap between in-domain and out-of-domain performance. They maintain high evaluation scores even when tested on previously unseen tumor types, especially models such as Virchow2 and H-optimus-0 where cross-domain results are very similar in every scenario. This robustness is critical for clinical deployment, where models should be able to generalize to new data distributions and avoid overfitting to specific domains. 

Adapting foundation models with parameter-efficient methods like \ac{lora} offers a strong compute–performance trade-off. \ac{lora} updates a small fraction of weights (typically <5\%), significantly reducing memory and training time compared to full fine-tuning. While linear probing is most efficient, \ac{lora} provides substantial performance gains with only modest additional compute. This makes adapted foundation models a pragmatic choice over training large Vision Transformers end-to-end, especially when compute, time, or data is limited. They offer strong accuracy with reduced training costs and deployment complexity. Rigorous quantification of these efficiencies is a key area for future research.

While we freeze or adapt transformer layers uniformly in our experiments, the need for adaptation is likely heterogeneous across depth. Early blocks tend to encode low-level features (e.g., edges, textures), whereas deeper blocks capture higher-level, task- and domain-specific concepts. Under domain shift, both strata can degrade. Early blocks may require modest recalibration when low-level statistics shift (e.g., noise, color, resolution), while later blocks may demand stronger adaptation to realign semantic representations. This motivates a more in-depth investigation into feature discrimination across model depth for future work \citep{Ammeling2025tmi}. 

Despite the promising results, several limitations should be acknowledged. First, while our experiments cover a range of diverse foundation models and common adaptation strategies, the scope of this benchmark study is limited to mitotic figure classification. The generalizability to other specific histopathological tasks, remains to be established. Second, mitotic figure classification is often performed as a secondary stage after detecting initial candidate objects through a detection or segmentation pipeline, hence the integration into a full mitotic figure detection pipeline and its evaluation needs further work for full integration into clinical practice. Third, while \ac{lora} proved highly effective, we did not exhaustively explore the effects of the hyperparameter or other parameter-efficient fine-tuning methods or combinations thereof, which could yield further improvements. 

The ResNet50 end-to-end baseline performed strongly, sometimes outperforming foundation models with linear probing or \ac{lora} adaptation. This suggests that, in some cases, traditional architectures with full fine-tuning can still be competitive, especially when sufficient labeled data is available. Future work should further investigate the conditions under which foundation models provide the greatest benefits over conventional approaches.

\section{Conclusion}

Our findings support the growing consensus that foundation models, when paired with efficient adaptation strategies, are poised to transform computational pathology by enabling robust, scalable solutions that generalize across tasks and domains. The demonstrated data efficiency and cross-domain robustness of \ac{lora}-adapted models are particularly relevant for clinical translation, where data heterogeneity and annotation scarcity are persistent challenges. Future research should extend these benchmarks to additional tasks, datasets, and adaptation methods, and explore strategies for further improving out-of-domain generalization.




\acks{This work was supported by the Bavarian State Ministry of Science and the Arts (project Fokus-TML) and by the German Research Foundation (DFG) (projects 520330054, 460333672 CRC1540 EBM).}

%
\ethics{The work follows appropriate ethical standards in conducting research and writing the manuscript, following all applicable laws and regulations regarding treatment of animals or human subjects.}

\coi{We declare we do not have conflicts of interest.}

\data{The CCMCT dataset can be found at \url{https://doi.org/10.6084/m9.figshare.c.4552445.v1}. The MIDOG 2022 dataset can be found at \url{https://zenodo.org/records/6547151}. The code can be found at \url{https://github.com/DeepMicroscopy/FoundationModelComparison}.}

\bibliography{export}


\clearpage
\appendix



\section{Additional Results from Dataset Scaling Experiments}


\begin{table}[!ht]
    \centering
    \caption{Results at 0.1\% dataset size of CCMCT dataset.}
    \label{tab:CCMCT_01_datascale}
    \resizebox{\columnwidth}{!}{%
         \begin{tabular}{lccc}
    \toprule
    Model & AUROC & Balanced ACC & Weighted F1 \\
    \midrule
    ResNet50 (End-To-End) & 0.69±0.03 & 0.63±0.03 & 0.59±0.07 \\
    ViT-S (End-To-End) & 0.68±0.03 & 0.62±0.03 & 0.61±0.07 \\
    ViT-B (End-To-End) & 0.61±0.02 & 0.58±0.02 & 0.56±0.07 \\
    ViT-H (End-To-End) & 0.62±0.03 & 0.58±0.02 & 0.57±0.02 \\    \midrule
    ResNet50 (LinProb) & 0.57±0.04 & 0.54±0.02 & 0.58±0.02 \\
    ViT-H (LinProb) & 0.57±0.03 & 0.55±0.02 & 0.58±0.03 \\    \midrule
    ViT-S DINOv3 (LinProb) & 0.58±0.04 & 0.55±0.03 & 0.58±0.03 \\
    Phikon (LinProb) & 0.61±0.03 & 0.58±0.03 & 0.61±0.03 \\
    UNI (LinProb) & 0.64±0.02 & 0.60±0.02 & 0.62±0.02 \\
    Virchow (LinProb) & 0.62±0.03 & 0.58±0.02 & 0.61±0.02 \\
    Virchow2 (LinProb) & 0.62±0.03 & 0.58±0.02 & 0.61±0.03 \\
    Prov-GigaPath (LinProb) & 0.65±0.02 & 0.61±0.02 & 0.63±0.01 \\
    H-optimus-0 (LinProb) & 0.63±0.03 & 0.60±0.02 & 0.62±0.03 \\    \midrule
    ViT-S DINOv3 (LoRA) & 0.60±0.02 & 0.57±0.02 & 0.54±0.05 \\
    Phikon (LoRA) & 0.72±0.04 & 0.65±0.04 & 0.65±0.07 \\
    UNI (LoRA) & 0.70±0.04 & 0.64±0.03 & 0.60±0.08 \\
    Virchow (LoRA) & 0.73±0.06 & 0.66±0.05 & 0.66±0.09 \\
    Virchow2 (LoRA) & 0.75±0.03 & 0.69±0.03 & 0.69±0.03 \\    
    Prov-GigaPath (LoRA) & 0.74±0.03 & 0.68±0.03 & 0.66±0.05 \\
    H-optimus-0 (LoRA) & 0.71±0.04 & 0.65±0.03 & 0.65±0.05 \\
    \bottomrule
    \end{tabular}%
        }
\end{table}

\begin{table}[!ht]
    \centering
    \caption{Results at 1\% dataset size of CCMCT dataset.}
    \label{tab:CCMCT_1_datascale}
    \resizebox{\columnwidth}{!}{%
         \begin{tabular}{lccc}
    \toprule
    Model & AUROC & Balanced ACC & Weighted F1 \\
    \midrule
    ResNet50 (End-To-End) & 0.75±0.02 & 0.68±0.02 & 0.70±0.02 \\
    ViT-S (End-To-End) & 0.74±0.01 & 0.67±0.01 & 0.67±0.03 \\
    ViT-B (End-To-End) & 0.67±0.06 & 0.62±0.05 & 0.60±0.07 \\
    ViT-H (End-To-End) & 0.68±0.03 & 0.62±0.03 & 0.61±0.03 \\    \midrule
    ResNet50 (LinProb) & 0.58±0.01 & 0.55±0.01 & 0.58±0.01 \\
    ViT-H (LinProb) & 0.65±0.01 & 0.60±0.01 & 0.63±0.01 \\    \midrule
    ViT-S DINOv3 (LinProb) & 0.65±0.01 & 0.59±0.01 & 0.62±0.01 \\
    Phikon (LinProb) & 0.67±0.01 & 0.62±0.01 & 0.64±0.01 \\
    UNI (LinProb) & 0.69±0.02 & 0.64±0.01 & 0.66±0.01 \\
    Virchow (LinProb) & 0.70±0.01 & 0.64±0.01 & 0.66±0.01 \\
    Virchow2 (LinProb) & 0.69±0.01 & 0.64±0.01 & 0.66±0.01 \\
    Prov-GigaPath (LinProb) & 0.71±0.01 & 0.65±0.01 & 0.67±0.01 \\
    H-optimus-0 (LinProb) & 0.70±0.01 & 0.64±0.01 & 0.67±0.01 \\    \midrule
    ViT-S DINOv3 (LoRA) & 0.64±0.07 & 0.59±0.06 & 0.55±0.13 \\
    Phikon (LoRA) & 0.79±0.07 & 0.72±0.05 & 0.73±0.04 \\
    UNI (LoRA) & 0.76±0.07 & 0.69±0.06 & 0.70±0.05 \\
    Virchow (LoRA) & 0.80±0.08 & 0.71±0.07 & 0.72±0.10 \\
    Virchow2 (LoRA) & 0.82±0.01 & 0.75±0.01 & 0.76±0.00 \\
    Prov-GigaPath (LoRA) & 0.82±0.00 & 0.74±0.02 & 0.75±0.02 \\
    H-optimus-0 (LoRA) & 0.80±0.07 & 0.73±0.06 & 0.74±0.06 \\
    \bottomrule
    \end{tabular}%
            }
\end{table}

\begin{table}[!ht]
    \centering
    \caption{Results at 10\% dataset size of CCMCT dataset.}
    \label{tab:CCMCT_10_datascale}
    \resizebox{\columnwidth}{!}{%
         \begin{tabular}{lccc}
    \toprule
    Model & AUROC & Balanced ACC & Weighted F1 \\
    \midrule
    ResNet50 (End-To-End) & 0.84±0.02 & 0.75±0.02 & 0.75±0.03 \\
    ViT-S (End-To-End) & 0.81±0.02 & 0.70±0.03 & 0.66±0.07 \\
    ViT-B (End-To-End) & 0.71±0.09 & 0.63±0.07 & 0.61±0.04 \\
    ViT-H (End-To-End) & 0.75±0.04 & 0.67±0.06 & 0.62±0.13 \\    \midrule
    ResNet50 (LinProb) & 0.67±0.00 & 0.61±0.00 & 0.64±0.00 \\
    ViT-H (LinProb) & 0.71±0.00 & 0.62±0.01 & 0.65±0.01 \\    \midrule
    ViT-S DINOv3 (LinProb) & 0.68±0.00 & 0.57±0.01 & 0.60±0.01 \\
    Phikon (LinProb) & 0.78±0.00 & 0.69±0.00 & 0.72±0.00 \\
    UNI (LinProb) & 0.78±0.00 & 0.70±0.00 & 0.72±0.00 \\
    Virchow (LinProb) & 0.79±0.00 & 0.71±0.00 & 0.73±0.00 \\
    Virchow2 (LinProb) & 0.78±0.00 & 0.70±0.00 & 0.72±0.00 \\
    Prov-GigaPath (LinProb) & 0.78±0.01 & 0.70±0.00 & 0.73±0.00 \\
    H-optimus-0 (LinProb) & 0.79±0.00 & 0.71±0.01 & 0.74±0.01 \\    \midrule
    ViT-S DINOv3 (LoRA) & 0.74±0.12 & 0.67±0.09 & 0.67±0.11 \\
    Phikon (LoRA) & 0.77±0.08 & 0.70±0.07 & 0.69±0.08 \\
    UNI (LoRA) & 0.82±0.06 & 0.74±0.05 & 0.74±0.06 \\
    Virchow (LoRA) & 0.84±0.06 & 0.76±0.05 & 0.76±0.05 \\
    Virchow2 (LoRA) & 0.85±0.03 & 0.76±0.03 & 0.78±0.02 \\
    Prov-GigaPath (LoRA) & 0.86±0.01 & 0.77±0.01 & 0.78±0.01 \\
    H-optimus-0 (LoRA) & 0.87±0.02 & 0.78±0.03 & 0.79±0.02 \\
    \bottomrule
    \end{tabular}%
            }
\end{table}

\begin{table}[!ht]
    \centering
    \caption{Results at 100\% dataset size of CCMCT dataset.}
    \label{tab:CCMCT_100_datascale}
    \resizebox{\columnwidth}{!}{%
         \begin{tabular}{lccc}
    \toprule
    Model & AUROC & Balanced ACC & Weighted F1 \\
    \midrule
    ResNet50 (End-To-End) & 0.84±0.02 & 0.75±0.02 & 0.75±0.03 \\
    ViT-S (End-To-End) & 0.81±0.02 & 0.70±0.03 & 0.66±0.07 \\
    ViT-B (End-To-End) & 0.71±0.09 & 0.63±0.07 & 0.61±0.04 \\
    ViT-H (End-To-End) & 0.75±0.04 & 0.67±0.06 & 0.62±0.13 \\    \midrule
    ResNet50 (LinProb) & 0.67±0.00 & 0.61±0.00 & 0.64±0.00 \\
    ViT-H (LinProb) & 0.71±0.00 & 0.62±0.01 & 0.65±0.01 \\    \midrule
    ViT-S DINOv3 (LinProb) & 0.68±0.00 & 0.57±0.01 & 0.60±0.01 \\
    Phikon (LinProb) & 0.78±0.00 & 0.69±0.00 & 0.72±0.00 \\
    UNI (LinProb) & 0.78±0.00 & 0.70±0.00 & 0.72±0.00 \\
    Virchow (LinProb) & 0.79±0.00 & 0.71±0.00 & 0.73±0.00 \\
    Virchow2 (LinProb) & 0.78±0.00 & 0.70±0.00 & 0.72±0.00 \\
    Prov-GigaPath (LinProb) & 0.78±0.01 & 0.70±0.00 & 0.73±0.00 \\
    H-optimus-0 (LinProb) & 0.79±0.00 & 0.71±0.01 & 0.74±0.01 \\    \midrule
    ViT-S DINOv3 (LoRA) & 0.74±0.12 & 0.67±0.09 & 0.67±0.11 \\
    Phikon (LoRA) & 0.77±0.08 & 0.70±0.07 & 0.69±0.08 \\
    UNI (LoRA) & 0.82±0.06 & 0.74±0.05 & 0.74±0.06 \\
    Virchow (LoRA) & 0.84±0.06 & 0.76±0.05 & 0.76±0.05 \\
    Virchow2 (LoRA) & 0.85±0.03 & 0.76±0.03 & 0.78±0.02 \\
    Prov-GigaPath (LoRA) & 0.86±0.01 & 0.77±0.01 & 0.78±0.01 \\
    H-optimus-0 (LoRA) & 0.87±0.02 & 0.78±0.03 & 0.79±0.02 \\
    \bottomrule
    \end{tabular}%
            }
\end{table}


\begin{table}[!ht]
    \centering
    \caption{Results at 0.1\% dataset size of MIDOG dataset.}
    \label{tab:MIDOG_01_datascale}
    \resizebox{\columnwidth}{!}{%
         \begin{tabular}{lccc}
    \toprule
    Model & AUROC & Balanced ACC & Weighted F1 \\
    \midrule
    ResNet50 (End-To-End) & 0.69±0.03 & 0.63±0.03 & 0.59±0.07 \\
    ViT-S (End-To-End) & 0.68±0.03 & 0.62±0.03 & 0.61±0.07 \\
    ViT-B (End-To-End) & 0.61±0.02 & 0.58±0.02 & 0.56±0.07 \\
    ViT-H (End-To-End) & 0.59±0.07 & 0.56±0.04 & 0.51±0.09 \\    \midrule
    ResNet50 (LinProb) & 0.57±0.04 & 0.54±0.02 & 0.58±0.02 \\
    ViT-H (LinProb) & 0.57±0.03 & 0.55±0.02 & 0.58±0.03 \\    \midrule
    ViT-S DINOv3 (LinProb) & 0.58±0.04 & 0.55±0.03 & 0.58±0.03 \\
    Phikon (LinProb) & 0.61±0.03 & 0.58±0.03 & 0.61±0.03 \\
    UNI (LinProb) & 0.64±0.02 & 0.60±0.02 & 0.62±0.02 \\
    Virchow (LinProb) & 0.62±0.03 & 0.58±0.02 & 0.61±0.02 \\
    Virchow2 (LinProb) & 0.62±0.03 & 0.58±0.02 & 0.61±0.03 \\
    Prov-GigaPath (LinProb) & 0.65±0.02 & 0.61±0.02 & 0.63±0.01 \\
    H-optimus-0 (LinProb) & 0.63±0.03 & 0.60±0.02 & 0.62±0.03 \\    \midrule
    ViT-S DINOv3 (LoRA) & 0.60±0.02 & 0.57±0.02 & 0.54±0.05 \\
    Phikon (LoRA) & 0.64±0.11 & 0.58±0.08 & 0.54±0.13 \\
    UNI (LoRA) & 0.63±0.08 & 0.57±0.07 & 0.50±0.12 \\
    Virchow (LoRA) & 0.67±0.14 & 0.60±0.08 & 0.56±0.14 \\
    Virchow2 (LoRA) & 0.67±0.11 & 0.60±0.10 & 0.55±0.16 \\
    Prov-GigaPath (LoRA) & 0.65±0.12 & 0.59±0.09 & 0.52±0.15 \\
    H-optimus-0 (LoRA) & 0.65±0.12 & 0.58±0.08 & 0.52±0.14 \\
    \bottomrule
    \end{tabular}%
            }
\end{table}

\begin{table}[!ht]
    \centering
    \caption{Results at 1\% dataset size of MIDOG dataset.}
    \label{tab:MIDOG_1_datascale}
    \resizebox{\columnwidth}{!}{%
         \begin{tabular}{lccc}
    \toprule
    Model & AUROC & Balanced ACC & Weighted F1 \\
    \midrule
    ResNet50 (End-To-End) & 0.75±0.02 & 0.68±0.02 & 0.70±0.02 \\
    ViT-S (End-To-End) & 0.74±0.01 & 0.67±0.01 & 0.67±0.03 \\
    ViT-B (End-To-End) & 0.67±0.06 & 0.62±0.05 & 0.60±0.07 \\
    ViT-H (End-To-End) & 0.70±0.03 & 0.63±0.03 & 0.61±0.04 \\    \midrule
    ResNet50 (LinProb) & 0.58±0.01 & 0.55±0.01 & 0.58±0.01 \\
    ViT-H (LinProb) & 0.65±0.01 & 0.60±0.01 & 0.63±0.01 \\    \midrule
    ViT-S DINOv3 (LinProb) & 0.65±0.01 & 0.59±0.01 & 0.62±0.01 \\
    Phikon (LinProb) & 0.67±0.01 & 0.62±0.01 & 0.64±0.01 \\
    UNI (LinProb) & 0.69±0.02 & 0.64±0.01 & 0.66±0.01 \\
    Virchow (LinProb) & 0.70±0.01 & 0.64±0.01 & 0.66±0.01 \\
    Virchow2 (LinProb) & 0.69±0.01 & 0.64±0.01 & 0.66±0.01 \\
    Prov-GigaPath (LinProb) & 0.71±0.01 & 0.65±0.01 & 0.67±0.01 \\
    H-optimus-0 (LinProb) & 0.70±0.01 & 0.64±0.01 & 0.67±0.01 \\    \midrule
    ViT-S DINOv3 (LoRA) & 0.64±0.07 & 0.59±0.06 & 0.55±0.13 \\
    Phikon (LoRA) & 0.75±0.10 & 0.67±0.10 & 0.66±0.13 \\
    UNI (LoRA) & 0.75±0.08 & 0.66±0.08 & 0.65±0.11 \\
    Virchow (LoRA) & 0.81±0.08 & 0.72±0.09 & 0.72±0.11 \\
    Virchow2 (LoRA) & 0.82±0.05 & 0.74±0.08 & 0.73±0.11 \\
    Prov-GigaPath (LoRA) & 0.80±0.06 & 0.70±0.08 & 0.70±0.11 \\
    H-optimus-0 (LoRA) & 0.82±0.07 & 0.73±0.08 & 0.73±0.10 \\
    \bottomrule
    \end{tabular}%
            }
\end{table}

\begin{table}[!ht]
    \centering
    \caption{Results at 10\% dataset size of MIDOG dataset.}
    \label{tab:MIDOG_10_datascale}
    \resizebox{\columnwidth}{!}{%
         \begin{tabular}{lccc}
    \toprule
    Model & AUROC & Balanced ACC & Weighted F1 \\
    \midrule
    ResNet50 (End-To-End) & 0.84±0.02 & 0.75±0.02 & 0.75±0.03 \\
    ViT-S (End-To-End) & 0.81±0.02 & 0.70±0.03 & 0.66±0.07 \\
    ViT-B (End-To-End) & 0.71±0.09 & 0.63±0.07 & 0.61±0.04 \\
    ViT-H (End-To-End) & 0.76±0.03 & 0.67±0.05 & 0.65±0.09 \\    \midrule
    ResNet50 (LinProb) & 0.67±0.00 & 0.61±0.00 & 0.64±0.00 \\
    ViT-H (LinProb) & 0.71±0.00 & 0.62±0.01 & 0.65±0.01 \\    \midrule
    ViT-S DINOv3 (LinProb) & 0.68±0.00 & 0.57±0.01 & 0.60±0.01 \\
    Phikon (LinProb) & 0.78±0.00 & 0.69±0.00 & 0.72±0.00 \\
    UNI (LinProb) & 0.78±0.00 & 0.70±0.00 & 0.72±0.00 \\
    Virchow (LinProb) & 0.79±0.00 & 0.71±0.00 & 0.73±0.00 \\
    Virchow2 (LinProb) & 0.78±0.00 & 0.70±0.00 & 0.72±0.00 \\
    Prov-GigaPath (LinProb) & 0.78±0.01 & 0.70±0.00 & 0.73±0.00 \\
    H-optimus-0 (LinProb) & 0.79±0.00 & 0.71±0.01 & 0.74±0.01 \\    \midrule
    ViT-S DINOv3 (LoRA) & 0.74±0.12 & 0.67±0.09 & 0.67±0.11 \\
    Phikon (LoRA) & 0.79±0.10 & 0.72±0.09 & 0.71±0.09 \\
    UNI (LoRA) & 0.84±0.05 & 0.76±0.04 & 0.76±0.04 \\
    Virchow (LoRA) & 0.81±0.11 & 0.73±0.10 & 0.74±0.11 \\
    Virchow2 (LoRA) & 0.87±0.03 & 0.79±0.03 & 0.80±0.02 \\
    Prov-GigaPath (LoRA) & 0.83±0.08 & 0.75±0.07 & 0.76±0.07 \\
    H-optimus-0 (LoRA) & 0.88±0.02 & 0.80±0.03 & 0.81±0.02 \\
    \bottomrule
    \end{tabular}%
            }
\end{table}

\begin{table}[!ht]
    \centering
    \caption{Results at 100\% dataset size of MIDOG dataset.}
    \label{tab:MIDOG_100_datascale}
    \resizebox{\columnwidth}{!}{%
         \begin{tabular}{lccc}
    \toprule
    Model & AUROC & Balanced ACC & Weighted F1 \\
    \midrule
    ResNet50 (End-To-End) & 0.87±0.01 & 0.79±0.01 & 0.78±0.01 \\
    ViT-S (End-To-End) & 0.82±0.01 & 0.73±0.01 & 0.72±0.03 \\
    ViT-B (End-To-End) & 0.79±0.01 & 0.70±0.02 & 0.71±0.03 \\
    ViT-H (End-To-End) & 0.80±0.04 & 0.70±0.05 & 0.70±0.09 \\    \midrule
    ResNet50 (LinProb) & 0.69±0.00 & 0.61±0.00 & 0.65±0.00 \\
    ViT-H (LinProb) & 0.71±0.00 & 0.61±0.00 & 0.65±0.00 \\    \midrule
    ViT-S DINOv3 (LinProb) & 0.68±0.00 & 0.57±0.00 & 0.60±0.00 \\
    Phikon (LinProb) & 0.81±0.00 & 0.71±0.00 & 0.74±0.00 \\
    UNI (LinProb) & 0.83±0.00 & 0.73±0.00 & 0.76±0.00 \\
    Virchow (LinProb) & 0.84±0.00 & 0.75±0.00 & 0.78±0.00 \\
    Virchow2 (LinProb) & 0.85±0.00 & 0.76±0.00 & 0.78±0.00 \\
    Prov-GigaPath (LinProb) & 0.83±0.00 & 0.74±0.00 & 0.77±0.00 \\
    H-optimus-0 (LinProb) & 0.84±0.00 & 0.75±0.00 & 0.78±0.00 \\    \midrule
    ViT-S DINOv3 (LoRA) & 0.78±0.12 & 0.70±0.09 & 0.71±0.09 \\
    Phikon (LoRA) & 0.80±0.13 & 0.73±0.10 & 0.73±0.11 \\
    UNI (LoRA) & 0.84±0.09 & 0.76±0.07 & 0.76±0.07 \\
    Virchow (LoRA) & 0.87±0.07 & 0.78±0.06 & 0.79±0.05 \\
    Virchow2 (LoRA) & 0.89±0.01 & 0.80±0.02 & 0.81±0.01 \\
    Prov-GigaPath (LoRA) & 0.87±0.05 & 0.79±0.04 & 0.79±0.04 \\
    H-optimus-0 (LoRA) & 0.90±0.02 & 0.81±0.02 & 0.81±0.02 \\
    \bottomrule
    \end{tabular}%
    }
\end{table}

\end{document}